\documentstyle[fleqn,br]{article}

\newcommand{\implies}{\supset}     

\newcommand{\theory}[1]{Cn({#1})}

\newcommand{\Th}{{\cal T}}

\newcommand{\Mod}[1]{{\mathit{Mod}\/({#1})}}

\renewcommand{\implies}{\supset}

\newtheorem{definition}{Definition}
\newtheorem{theorem}{Theorem}

\newcommand{\revise}{\dot{+}}
\newcommand{\reviseC}{\revise_{\!c}} 
\newcommand{\contract}{\dot{-}}
\newcommand{\contractC}{\contract_{\!c}} 

\newcommand{\update}{\diamond}

\title{A Consistency-Based Model for Belief Change: Preliminary Report}
\author{James P.\ Delgrande\\
  School of Computing Science,\\
  Simon Fraser University \\
  Burnaby, B.C.,
  Canada  V5A 1S6 \\
  jim@cs.sfu.ca
  \And
  Torsten Schaub \\
  Institut f\"ur Informatik,\\
  Universit\"at Potsdam      \\  
  D--14415 Potsdam,
  Germany\\
  torsten@cs.uni-potsdam.de}
\date{}
\begin{document}
\maketitle

\begin{abstract}
We present a general, consistency-based framework for belief change.
Informally, in revising $K$ by $\alpha$, we begin with $\alpha$ and
incorporate as much of $K$ as consistently possible.
Formally, a knowledge base $K$ and sentence $\alpha$ are expressed, via
renaming propositions in $K$, in separate alphabets, but such that there is
an isomorphism between the original and new alphabets.
Using a maximization process, we assume that corresponding atoms in each
language are equivalent insofar as is consistently possible.
Lastly, we express the resultant knowledge base using just the original
alphabet.
There may be more than one way in which $\alpha$ can be so extended by $K$:
in {\em choice revision}, one such ``extension'' represents the revised
state;
alternately {\em revision} consists of the intersection of all such
extensions.

The overall framework is flexible enough to express
other approaches to revision and update,
and the incorporation of static and dynamic integrity constraints.
Our framework differs from work based on ordinal conditional functions,
notably with respect to iterated revision.
We argue that the approach is well-suited for implementation:
choice revision gives better complexity results than general revision;
the approach can be expressed in terms of a finite knowledge base; and
the scope of a revision can be restricted to just those propositions
mentioned in the sentence for revision $\alpha$.
\end{abstract}


\section{Introduction}
\label{sec:introduction}

We describe a general framework for belief change.
The approach has something of the same flavour as the consistency-based
paradigm for diagnosis \cite{Reiter87b} or the assumption-based approach
to default reasoning \cite{Poole88}, although it differs significantly in
details.
Informally, in revising a knowledge base $K$ by sentence $\alpha$, we
begin with $\alpha$ and incorporate as much of $K$ as consistently
possible.
There may be more than one way in which information from $K$ can be
incorporated.
This gives rise to two notions of revision:
a choice notion, in which one such ``extension'' is used for the revised
state, and the intersection of all such extensions.
Belief contraction is defined analogously.

We mainly focus on belief revision in this paper.
For revision, first a knowledge base $K$ and sentence $\alpha$ are
expressed, via renaming atomic propositions in $K$, in separate alphabets.
We next assume that as many atoms in $\alpha$ are equivalent to the
corresponding atom in $K$, as consistently possible.
A set of such equivalent atoms is used to incorporate as much of the
original knowledge base as is consistently possible.
In the final section we discuss the more general approach,
which we show is flexible enough to express extant approaches to revision and
update.

The approach is developed in a formal, abstract framework.
However, we argue that it is well-suited for implementation:
The notion of choice revision gives better complexity results than
general revision;
moreover, we argue that belief revision is an area in which choice reasoning
makes sense in some cases.
Second, we show how the approach can be expressed equivalently in terms of a
finite knowledge base, in place of a deductively-closed belief set.
Third, we show that the scope of a revision can be restricted to just those
propositions common to the knowledge base and sentence for revision.

We begin by presenting a very general framework for expressing belief change.
This is restricted to address revision and contraction.
Following this, we show how the approach allows for a uniform treatment of
integrity constraints.
As well, the approach supports iterated revision, with properties distinct
from approaches based on the work of Spohn \cite{Spohn88}.
Finally we briefly explore the general framework, and suggest it is flexible
enough to express extant approaches to revision and update.

\section{Background}
\label{sec:background}

A common approach in belief revision is to provide a set of
{\em rationality postulates} for revision and contraction functions.
The {\em AGM approach} of Alchourron, G{\"{a}}rdenfors, and Makinson
\cite{Gardenfors88}, provides the best-known set of such postulates.
The goal is to describe belief change
on an abstract level, independent of how beliefs are represented and
manipulated.
Belief states, called {\em belief sets}, are modelled by sets of sentences
closed under the logical consequence operator of some logic in some
language ${\cal L}$, where the logic includes classical propositional logic.
For belief set $K$,
$K + \alpha$ is the deductive closure of $K \cup \{\alpha\}$,
and is called the {\em expansion} of $K$ by $\alpha$.
$K_\bot$ is the inconsistent belief set (i.e.\ $K_\bot = {\cal L}$).
$\Th$ is the set of all belief sets.

A {\em revision} function $\revise$ is a function from
$\Th\times {\cal L}$ to $\Th$ satisfying the following postulates.
\begin{description}
\item[($K \revise 1$)]
$K \revise \alpha$ is a belief set.

\item[($K \revise 2$)]
\(
\alpha \in K \revise \alpha.
\)

\item[($K \revise 3$)]
\(
K \revise \alpha \subseteq K + \alpha.
\)

\item[($K \revise 4$)]
If
\(
\neg \alpha \not\in K,
\)
then
\(
K + \alpha \subseteq K \revise \alpha.
\)

\item[($K \revise 5$)]
\(
K \revise \alpha = K_\bot
\)
iff
\(
\vdash \neg \alpha.
\)

\item[($K \revise 6$)]
If
\(
\vdash
\alpha \equiv \beta,
\)
then
\(
K \revise \alpha = K \revise \beta.
\)

\item[($K \revise 7$)]
\(
K \revise (\alpha\wedge\beta) \subseteq (K \revise \alpha) + \beta.
\)

\item[($K \revise 8$)]
If
\(
\neg \beta \not\in K \revise \alpha,
\)
then
\(
(K \revise \alpha) + \beta \subseteq K \revise (\alpha \wedge \beta).
\)
\end{description}
That is:
the result of revising $K$ by $\alpha$ is a belief set in which $\alpha$
is believed; 
whenever the result is consistent, revision consists of the expansion of
$K$ by $\alpha$;
the only time that $K_\bot$ is obtained is when $\alpha$ is
inconsistent;
and revision is independent of the syntactic form of $K$ and $\alpha$.
The last two postulates deal with the relation between revising with a
conjunction and expansion.

\cite{KatsunoMendelzon92} explores the distinct notion of belief {\em update}
in which an agent changes its beliefs in response to changes in its
external environment.
Our interests here centre on revision;
however as the end of the paper, we briefly consider this approach.

Recently there has been interest in {\em iterated} belief revision,
a topic that the AGM approach by-and-large leaves open.
Representative work includes
\cite{Boutilier94a,Williams94,Lehmann95,DarwichePearl97}.
We discuss Darwiche and Pearl's approach here.
They employ the notion of an {\em epistemic state} that encodes how the
revision function changes following a revision.
$\Psi$ denotes an epistemic state;
$Bel(\Psi)$ denotes the belief set corresponding to $\Psi$.
So now the result of revising an epistemic state is another epistemic state
(from which the revised belief set may be determined using $Bel(\cdot)$.
Darwiche and Pearl propose the following postulates that ``{\em any} rational
system of belief change should comply with'' (p.\ 2).
Following their practice, we use $\Psi$ to stand for $Bel(\Psi)$ when it
appears as an argument of $\models$.

\begin{description}
\item[$C1$:]
If
$\alpha \models \mu$
then
$(\Psi \revise \mu)\revise \alpha \equiv \Psi \revise \alpha$.
\item[$C2$:]
If
$\alpha \models \neg \mu$
then
$(\Psi \revise \mu)\revise \alpha \equiv \Psi \revise \alpha$.
\item[$C3$:]
If
$\Psi \revise \alpha \models \mu$
then
$(\Psi \revise \mu)\revise \alpha \models \mu$.
\item[$C4$:]
If
$\Psi \revise \alpha \not \models \neg \mu$
then
$(\Psi \revise \mu)\revise \alpha \not \models \neg \mu$.
\end{description}
\cite{NayakFooPagnuccoSattar96} propose a variant of $C2$ along with the
following postulate:
\begin{description}
\item[$Conj$:]
If
$\alpha \wedge \beta \not\models \bot$
then
\(
(\Psi \revise \alpha) \revise^\alpha \beta
=
\Psi \revise (\alpha\wedge\beta).
\)
\end{description}
where $\revise^\alpha$ indicates that the change in $\revise$ following
revision by $\alpha$ depends in part on $\alpha$.
This postulate is shown to be strong enough to derive $C1$, $C3$, and $C4$ in
the presence of the other postulates.

There has also been work on specific approaches to revision based on the
distance between models of a knowledge base and a sentence to be incorporated
in the knowledge base.
This work includes 
\cite{Dalal88,Forbus89,Satoh88,Winslett88}.
In these approaches, models of the new knowledge base consist of models of
the sentence to be added that are closest (based on ``distance'' between
atomic sentences) to models of the original knowledge base.

Our approach differs from previous work first, in that we provide a
specific, albeit general, framework in which approaches may be expressed.
As well, the general framework allows the incorporation of different forms
of integrity constraints.
Also, given that it falls into the ``consistency-based'' paradigm, the
approach has a certain syntactic flavour.
However, notably, our approach is independent of the syntactic form of the
knowledge base and sentence for revision.

Our technique of maximizing sets of equivalences of propositional letters
bears a superficial resemblance to the use of such equivalences in
\cite{LiberatoreSchaerf97} (based in turn on the technique developed in
\cite{deKleerKonolige89}).
However the approaches are distinct;
in particular and in contradistinction to these references, we employ
disjoint alphabets for a knowledge base and revising sentence.
As well, the approach bears a resemblance to that of \cite{delVal93}.
However, unlike del Val, we provide a single approach which may be restricted
to yield extant approaches;
also, we place no a priori restrictions on the form of a knowledge base.

\section{Formal Preliminaries}
\label{sec:definitions}

We deal with propositional languages and use the logical symbols
$\top$, $\bot$, $\neg$, $\vee$, $\wedge$, $\implies$, and $\equiv$
to construct formulas in the standard way.
We write ${\cal L}_{{\cal P}}$ to denote a language over an alphabet
${\cal P}$
of {\em propositional letters} or {\em atomic propositions}.
Formulas are denoted by the Greek letters $\alpha$, $\beta$, $\alpha_1$,
\dots.
{\em Knowledge bases} or, equivalently, {\em belief sets} are initially
identified with deductively-closed sets of formulas and are denoted
$K$, $K_1$, \dots.
So we have $K = \theory{K}$, where $\theory{\cdot}$ is the deductive closure
of the formula or set of formulas given as argument.
Later we relax this restriction.

Given an alphabet ${\cal P}$, we define a disjoint alphabet
${\cal P}'$ as
\(
{\cal P}'=\{p'\mid p\in {\cal P}\}
\).
Then, for $\alpha \in {\cal L}_{{\cal P}}$, we define
$\alpha'$ as the result of replacing in $\alpha$ each proposition $p$ from
${\cal P}$ by the corresponding proposition $p'$ in ${\cal P}'$
(so implicitly there is an isomorphism between ${\cal P}$ and
${\cal P}'$).
This is defined analogously for sets of formulas.

We define a \emph{belief change scenario} in language
${\cal L}_{{\cal P}}$ as a triple $B = (K, U, V)$,
where $K,U,V$ are sets of formulas in ${\cal L}_{{\cal P}}$.
Informally, $K$ is a knowledge base that will be changed such that the set
$U$ will be true in the result, and the set $V$ will be consistent with the
result.
For a base approach to revision we take $V = \emptyset$ and for a base
approach to contraction we take $U = \emptyset$.

In the definition below, ``maximal'' is with respect to set containment
(rather than set cardinality).
The following is our central definition.
\begin{definition}\label{def:extension}
  Let $B=(K,U,V)$ be a belief change scenario in ${\cal L}_{{\cal P}}$.
  Define $EQ$ as a maximal set of equivalences
  \(
  EQ\subseteq\{p\equiv p'\mid p\in{\cal P}\}
  \)
  such that
  \[
  \quad\;\;\; K'\cup EQ\cup U\cup V \not\vdash \bot.
  \]
  Then
  \[
  \theory{K'\cup EQ\cup U}\cap{\cal L}_{{\cal P}}
  \]
  is a {\em consistent definitional extension} of $B$.
\end{definition}
Hence, a consistent definitional extension of $B$ is a modification of
$K$ in which $U$ is true, and in which $V$ is consistent.
We say that $EQ$ {\em underlies} the consistent definitional extension of
$B$.
We let $\overline{EQ}$ stand for
\(
\{p\equiv p'\mid p\in{\cal P}\}\setminus EQ
\).

Clearly, for a given belief change scenario there may be more than one
consistent definitional extension.
We will make use of the notion of a {\em selection function} $c$ that
for any set $I \neq \emptyset$ has as value some element of $I$.
In Definition~\ref{def:revision} and~\ref{def:contraction},
these primitive functions can be regarded as inducing selection functions $c'$
on belief change scenarios, such that $c'((K,U,V))$ has as value some consistent
definitional extension of $(K,U,V)$.
This is a slight generalisation of selection functions as found in the AGM
approach.
%
%
%
%
%

\section{Revision and Contraction}
\label{sec:revision.contraction}

Definition~\ref{def:extension} provides a very general framework for
specifying belief change.
In the next two definitions we give specific definitions for revision and
contraction.
We develop these specific approaches and then, at the end of the paper,
we return to the more general framework of Definition~\ref{def:extension}
and discuss how it can be used to express other approaches.

\begin{definition}[Revision]
\label{def:revision}
Let $K$ be a knowledge base and $\alpha$ a formula, 
and let $(E_i)_{i\in I}$ be the family of all 
consistent definitional extensions of
\(
(K, \{\alpha\}, \emptyset)
\).
Then
\begin{enumerate}
\item
\(
K\reviseC \alpha = E_i
\)
\quad
is a {\em choice revision} of $K$ by $\alpha$
with respect to some selection function $c$ with $c(I)=i$.

\item
\(
K\revise\alpha
= \bigcap_{i\in I} E_i
\) 
\quad
is the {\em (skeptical) revision} of $K$ by $\alpha$. 
\end{enumerate}
\end{definition}
Table~\ref{fig:revision.egs} gives examples of (skeptical) revision.
The first column gives the original knowledge base, but with atoms already
renamed.
The second column gives the revision formula,
while the third gives the $EQ$ set(s) 
and the last column gives the results of the revision.
For the first and last column, we give a formula whose deductive closure
gives the corresponding belief set.
\begin{table}[h]
\[
\hspace{-1.5em}
\begin{array}{c|c|c|c}
K' & \alpha & EQ & K \revise \alpha
\\
\hline
p' \wedge q' & \neg q & \{p \equiv p' \} & p \wedge \neg q
\\
\neg p'  \equiv q' & \neg q & \{\, p \equiv p', \; q\equiv q' \, \} & p \wedge \neg q
\\
p'  \vee q' & \neg p \vee\neg q & \{ \, p \equiv p', \; q\equiv q' \, \} & p \equiv\neg q
\\
p' \wedge q' & \neg p \vee\neg q & \{p \equiv p' \}, \;  \{q\equiv q'\} & p \equiv\neg q \\
\end{array}
\]
\caption{\label{fig:revision.egs}(Skeptical) revision examples.}
\end{table}

\noindent
In detail, for the last example, we wish to determine
\begin{equation}\label{eq:ex:rev:tri}
  \{p \wedge q\}\revise{(\neg p \vee\neg q)} \ .
\end{equation}
We find maximal sets 
\(
EQ
\subseteq
\{p \equiv p', q\equiv q'\}
\)
such that
\[
\{p' \wedge q'\}\cup EQ\cup\{\neg p \vee\neg q\}\cup\emptyset
\mbox{ is consistent.}
\]
We get two such sets of equivalences, namely
\(             
  EQ_1
  =
  \{p \equiv p' \}
\) and \(      
  EQ_2
  =
  \{q\equiv q'\}
\).            
Accordingly, we obtain
\[             
  \{p \wedge q\}\revise{(\neg p \vee\neg q)}
  =
\]
\[
\quad
  \mbox{$\bigcap_{i=1,2}$}
  \theory{\{p' \wedge q'\}\cup EQ_i\cup\{\neg p \vee\neg q\}}
  \cap{\cal L}_{{\cal P}}.
\]             
In addition to $(\neg p \vee\neg q)$, we get $(p \vee q)$,
jointly implying $(p \equiv\neg q)$.

In this example we get two choice extensions, $\theory{p \wedge \neg q}$ and
$\theory{\neg p \wedge q}$.
This raises the question of the usefulness of choice revision compared to
general revision.
An apparent limitation of a choice reasoner is that it might draw overly
strong conclusions.
However, in belief revision this may be less of a problem than, say, in
nonmonotonic reasoning:
the goal in revision is to determine the true state of the world;
if a (choice) revision results in an inaccurate knowledge base, then
{\em this} inaccuracy will presumably be detected and rectified in a later
revision.
So choice revision may do no worse than a ``skeptical'' operator with
respect to ``converging'' to the true state of the world.
In addition, as we later show, it may do so significantly more efficiently
and with better worst-case behaviour.
Hence for a land vehicle exploring a benign environment, choice revision
might be an effective part of a control mechanism;
for something like flight control, or controlling a nuclear reactor, one
would prefer skeptical revision.

Contraction is defined similarly to revision.
\begin{definition}[Contraction]
\label{def:contraction}
Let $K$ be a knowledge base and $\alpha$ a formula, 
and let $(E_i)_{i\in I}$ be the family of all 
consistent definitional extensions of
\(
(K,\emptyset, \{\neg\alpha\})
\).
Then 
\begin{enumerate}
\item
\(
K \contractC \alpha = E_i
\)
\quad
is a \/ {\em choice contraction} of $K$ by $\alpha$ 
with respect to some selection function $c$ with $c(I)=i$.

\item
\(
K\contract\alpha
= \bigcap_{i\in I} E_i
\)
\quad
is the {\em (skeptical) contraction} of $K$ by $\alpha$.
\end{enumerate}
\end{definition}
Table~\ref{fig:contraction.egs} gives examples of (skeptical) contraction,
using the same format and conventions as Table~\ref{fig:revision.egs}.
\begin{table}[h]
\[
\hspace{-1.5em}
\begin{array}{c|c|c|c}
K' & \alpha & EQ & K \contract \alpha
\\
\hline
p' \wedge q' & q & \{p \equiv p' \} & p
\\
p'  \wedge q' \wedge r' & p \vee q & \{r\equiv r'\} & r
\\
p'  \vee q' & p \wedge q & \{ \, p \equiv p', \; q\equiv q' \, \} & p \vee q
\\
p' \wedge q' & p \wedge q & \{p \equiv p' \}, \;  \{q\equiv q'\} & p \vee q
\\
\end{array}
\]
\caption{\label{fig:contraction.egs}(Skeptical) contraction examples.}
\end{table}

\noindent
In detail, for the first example we wish to determine
\begin{equation}
  \label{eq:ex:con:one}
  \{p \wedge q\}\contract{q} \ .
\end{equation}
We compute the consistent definitional extensions of
\(
(\{p \wedge q\},\emptyset,\{ \neg q\})
\).
We rename the propositions in $\{p \wedge q\}$ and look for maximal
subsets $EQ$ of
\(
\{p \equiv p', q\equiv q'\}
\)
such that
\[
\{p' \wedge q'\}\cup EQ\cup\emptyset\cup\{ \neg q\}
\mbox{ is consistent.}
\]
We obtain
\(
EQ=\{p \equiv p' \},
\)
yielding
\begin{eqnarray*}
\{p \wedge q\}\contract{q}
 & = &
  \theory{\{p' \wedge q'\}\cup\{p \equiv p' \}\cup\emptyset}\cap{\cal L}_{{\cal P}}
\\
 & = &
  \theory{\{p \}}.
\end{eqnarray*}

\subsection{Properties of Revision and Contraction}

With respect to the AGM postulates, we obtain the following.
\begin{theorem}
Let $\revise$ and $\reviseC$ be defined as in Definition~\ref{def:revision}.
Then $\revise$ and $\reviseC$ satisfy the basic AGM postulates
$(K \revise 1)$ to $(K \revise 4)$,
$(K \revise 6)$
as well as
$(K \revise 7)$.
\end{theorem}

For ($K \revise 5$) we have instead the weaker postulate:
\begin{description}
\item[($K \revise 5$)]
\(
K \revise \alpha = K_\bot
\)
iff:
\(
K = K_\bot
\mbox{ or }
\vdash \neg \alpha.
\)
\end{description}

We obtain analogous results for $\contract$ and $\contractC$ with respect
to the AGM contraction postulates:
\begin{theorem}
Let $\contract$ and $\contractC$ be defined as in
Definition~\ref{def:contraction}.
Then
$\contract\phantom{_c}$ satisfies the basic AGM postulates
$(K \contract 1)$ to $(K \contract 4)$,
$(K \contract 6)$, and $(K \contract 7)$.
%
In addition, $\contractC$ satisfies the basic AGM postulates
$(K \contract 1)$ to $(K \contract 4)$, $(K \contract 6)$.
\end{theorem}
%
%
%

\noindent
We also obtain the following interdefinability results:

\begin{theorem}[Levi Identity]
\label{thm:Levi}
  \(
  K \revise \alpha
  =
  (K \contract \neg\alpha) + \alpha.
  \)
\end{theorem}
\begin{theorem}[Partial Harper Identity]
\label{thm:Harper}
\ \\
\(
K \contract \alpha
\subseteq
K
\cap
(K \revise \neg\alpha)
\)
\end{theorem}
The following example shows that equality fails in the Harper Identity:
if $K \equiv p \wedge q \wedge r$ and $\alpha \equiv r$, then
\(
K \contract \alpha \equiv r
\)
while
\(
K
\cap
(K \revise \neg\alpha)
\equiv
(p \equiv \neg q) \wedge r.
\)
Similar results are obtained for choice revision and choice contraction by
appeal to appropriate selection functions.

\paragraph{Iterated belief change:}

The approach obviously supports iterated revision.
Since we use a ``global'' metric, and since we can assume that {\em every}
revision result, given $K$ and $\alpha$, can be determined, we continue to
use $K$ here rather than Darwiche and Pearl's $\Psi$ for an epistemic state.
That is, for us, we don't need to refer to epistemic states, since we have
completely specified how $\revise$ should behave on all arguments.
Nonetheless, neither operator in Definition~\ref{def:revision} satisfies
any of the Darwiche-Pearl postulates for iterated revision.
Nor in our opinion should they.
For example, for $C1$, if we have
\begin{equation}
\label{eq:C1}
K = \theory{\neg p}, \quad \alpha = p, \quad \mu = p \vee q,
\end{equation}
then in our approach we obtain that
\begin{equation}
\label{eq:C1b}
(K \revise \mu)\revise \alpha = \theory{p \wedge q}
\mbox{ but } K \revise \alpha = \theory{p}.
\end{equation}
\cite{DarwichePearl97,NayakFooPagnuccoSattar96}
assert that these results should be equal.
However, it is {\em possible} (contra $C1$) that there are cases where
revising $\neg p$ by $p \vee q$ yields $\neg p \wedge q$ and a subsequent
revision by $p$ then gives $p \wedge q$, but revising $\neg p$ by $p$ would
yield $p$.
Which is to say, a significant difficulty in the area of belief revision
is that different people have conflicting intuitions.
However, Darwiche and Pearl argue that {\em all rational} revision functions
should obey $C1$.
Consequently they would need to argue that in all cases, having
(\ref{eq:C1b}) result from (\ref{eq:C1}) is {\em irrational}.

More seriously,
an instance of $C2$ (letting $\alpha$ be $\neg \phi$ and $\mu$
be $\phi \wedge \psi$, whence $\alpha \models \neg \mu$) is the following:
\begin{description}
\item[$C2'$:]
$(K \revise (\phi \wedge \psi))\revise \neg \phi \equiv K \revise \neg \phi$.
\end{description}
Thus if you revise by ($\phi \wedge \psi$) and then revise by the negation
of some of this information ($\neg \phi$), then the other original
information ($\psi$) is lost.
So, in a variant of an example from \cite{DarwichePearl97}, consider where
I see a new bird in the distance and come
to believe that it is red and flies.
If on closer examination I see that it is yellow, then according to $C2'$ and
so $C2$, I also no longer believe that it flies.
This seems too strong a condition to want to adopt.
We conjecture (but have no proof) that approaches based on \cite{Spohn88},
such as \cite{DarwichePearl97}, are subject in some form to such a
``blanketing'' result.

On the other hand, there are nontrivial results concerning iterated revision
that hold for the present approach.
For example, we have:
\begin{theorem}
Let $\revise$ be defined as in Definition~\ref{def:revision}.
Then: 
\(
(\alpha \revise \beta) \revise \alpha
=
\beta \revise \alpha\ .
\)
\end{theorem}
%

\paragraph{Semantics:}

The operator
$\revise$ provides a (near) syntactic counterpart to the
minimal-distance-between-models approach of \cite{Satoh88}.
For two sets $S$ and $T$, let $S\Delta T$ be the symmetric difference,
\(
(S\cup T)\setminus (S\cap T)
\).
For formulas $\alpha,\beta$, define
\[
\Delta^{\min} (\alpha,\beta)
=
\]
\[
\quad
\mbox{$\min_\subseteq$}
      (\{M\Delta M'\mid M\in\Mod{\alpha},M'\in\Mod{\beta}\})\ ,
\]
where $\Mod{\alpha}$ is the set of all models of $\alpha$, each
of which is identified with a set of propositions.
Then, we have:
\begin{theorem}\label{thm:eq:model}
  Let $B=(K,U,\emptyset)$ be a belief change scenario in ${\cal L}_{{\cal P}}$
  where $K \neq {\cal L}_{{\cal P}}$,
  and let $(EQ_i)_{i\in I}$ be the family of all sets of equivalences,
  as defined in Definition~\ref{def:extension}.

  Then,
  \(
  \{\ \{p\in{\cal P}\mid (p\equiv p')\not\in{EQ_i}\}\ \mid i\in I\}
  =
  \Delta^{\min} (U,K).
  \) 
\end{theorem}
This correspondence is interesting, but is of limited use beyond supplying a
semantics for one instance of the approach.
The choice approach, and (below) considerations on implementation and
integrity constraints, are not readily expressed in Satoh's model-based
semantics.
As well, a contraction function is straightforwardly obtained in Satoh's
approach only by using the Harper Identity (which doesn't fully obtain
here).
Further, in the last section, we show how other approaches can
be expressed in our general framework.

\subsection{Integrity Constraints}

Definitions~\ref{def:revision}~and~\ref{def:contraction} are similar in
form, differing only in how the formula $\alpha$ is mapped onto the sets
$U$ and $V$ in Definition~\ref{def:extension}.
Clearly one can combine these definitions, allowing simultaneous
revision by one formula and contraction by another.
This in-and-of-itself isn't overly interesting, but it does lead to a
natural and general treatment of integrity constraints in our approach.

There are two standard definitions of a knowledge base $K$ satisfying a
static integrity constraint $IC$.
In the {\em consistency-based} approach of \cite{Kowalski78},
$K$ satisfies $IC$ iff $K \cup \{IC\}$ is satisfiable.
In the {\em entailment-based} approach of \cite{Reiter84},
$K$ satisfies $IC$ iff $K \vdash IC$.
\cite{KatsunoMendelzon91} show how entailment-based constraints can be
maintained across revisions:
given an integrity constraint $IC$
and revision function $\revise$,
a revision function $\revise^{IC}$
which preserves $IC$ is defined by:
\(
K \revise^{IC} \alpha
=
K \revise (\alpha \wedge IC).
\)
In our approach, we can define revision taking into account both approaches
to integrity constraints.

Corresponding to Definition~\ref{def:revision} (and ignoring the choice
approach) we obtain:
\begin{definition}
\label{def:revision.IC}
Let $K$ be a knowledge base, $\alpha$ a formula, and $IC_K$, $IC_R$ sets of
formulas.
Let $(E_i)_{i\in I}$ be the family of all
consistent definitional extensions of
\(
(K,\{\alpha\} \cup IC_R,IC_K)
\).
Then
\(
K\revise^{(IC_K,IC_R)} \alpha
= \bigcap_{i\in I} E_i
\)
\quad
is the {\em revision} of $K$ by $\alpha$ incorporating integrity constraints
$IC_K$ (consistency-based) and $IC_R$ (entailment-based).
\end{definition}
\begin{theorem}
Let $\revise^{(IC_K,IC_R)}$ be defined as in
Definition~\ref{def:revision.IC}.
%
Then
$\left(K \revise^{(IC_K,IC_R)} \alpha \right) \vdash IC_R$.
\quad
If $IC_R \cup IC_K \not\vdash \neg\alpha$ then
$\left(K \revise^{(IC_K,IC_R)} \alpha \right) \cup IC_K$ is satisfiable.
\end{theorem}

Finally, and in contrast with previous approaches, it is straightforward
to add {\em dynamic} integrity constraints, which express constraints that
hold between states of the knowledge base before and after revision.
The simplest way of so doing is to add such constraints to the set $V$ in
Definition~\ref{def:extension}.
To state that if $a \wedge b$ is true in a knowledge base before revision
then $c$ must be true afterwards, we would add $a' \wedge b' \implies c$ to
$V$.
Note however that the addition of dynamic constraints may lead to an operator
that violates some of the properties of $\revise$.
For example $\theory{\alpha} \revise \neg\alpha$ with dynamic constraint
$\alpha' \implies \alpha$ leads to an inconsistent revision.

\section{Implementability Considerations}
\label{sec:implementation}

We claimed at the outset that the approach is well-suited for implementation.
To this end, we first consider the use of choice belief revision.
Second we consider the problem of representing the results of revision
in a finite, manageable representation.
Lastly, we address limiting the range of $EQ$.

\paragraph{Complexity:}

From \cite{EiterGottlob92} and Theorem~\ref{thm:eq:model} it follows that
deciding, for given $K$, $\alpha$, $\beta$, whether
\(
K \revise \alpha \vdash \beta
\)
is
\(
\Pi^P_2\mbox{-complete}.
\)
However, the analogous problem for choice revision is lower in the
polynomial hierarchy.
%
\begin{theorem}\
Given a selection function $c$, formulas $K, \alpha, \beta$,
and a set of equivalnces $EQ$.
Then, we have:
\begin{enumerate}
\item
Deciding whether $EQ$ determines a choice revision of $K$ and $\alpha$
is in ${\bf P^{NP}}$.

\item
Deciding
\(
K \reviseC \alpha \vdash \beta
\)
is in ${\bf P^{NP}}$.
\end{enumerate}
\end{theorem}
%
We have not yet addressed restrictions on the syntactic form of $K$ or
$\alpha$;
but see \cite{EiterGottlob92}.

\paragraph{Finite representations:}

Definitions~\ref{def:extension},~\ref{def:revision},~and~\ref{def:contraction}
provide a characterisation of revision and contraction, yielding in either
case a deductively-closed belief set.
Here we consider how the same (with respect to logical equivalence)
operators can be defined, but where a knowledge base is given as an
arbitrary, finite set of formulas.
It follows from the discussion below that, for knowledge base
$K$ and formula $\alpha$, we can defined choice revision so that
$|K \reviseC \alpha| \leq |K| + |\alpha|$ for any selection function $c$.

Informally the procedure is straightforward, although the technical details
are less so.
A knowledge base $K$ is now represented by a formula (or set of formulas).
Via Definitions~\ref{def:extension}~and~\ref{def:revision} we consider
maximal sets $EQ$ where
\(
\{K'\} \cup \{\alpha\} \cup EQ
\)
is consistent.
For each such set $EQ$, we replace each $p'$ in $K'$ by $p$ where
$(p \equiv p') \in EQ$ and we replace each $p'$ in $K'$ by $\neg p$ where
$(p \equiv p') \in \overline{EQ}$.
The result of these substitutions into $\{K'\} \cup \{\alpha\}$ is a
sentence of size $\leq |K| + |\alpha|$ and whose deductive closure is
equivalent to (some) choice revision.
The disjunction of all such sentences (and so considering all possible sets
$EQ$) is equivalent to $\theory{K} \revise \alpha$.

As opposed to the computation of the sets $EQ$,
the result of revising or contracting a formula $K$ can be captured
without an explicit change of alphabet.
We start by observing that any set of equivalences
\(
EQ 
\)
induces a binary partition of its underlying alphabet ${\cal P}$,
namely
\(
\langle{\cal P}_{EQ},{\cal P}_{\overline{EQ}}\rangle
\)
with
\(
{\cal P}_{EQ}
=
\{p\in{\cal P}\mid p\equiv p'\in EQ\}
\)
and
\(
{\cal P}_{\overline{EQ}}
=
{\cal P}\setminus{\cal P}_{EQ}
\).
Given a belief change scenario $B$ along with a set of equivalences $EQ_i$
(according to Definition~\ref{def:extension}),
we define for
\(
\alpha
\in
{\cal L}_{{\cal P}}
\),
that
\(
\lceil\alpha\rceil_i
\)
is the result of replacing in $\alpha$ each proposition
\(
p\in{\cal P}_{\overline{EQ_i}}
\)
by its negation $\neg p$.

For generality, let $K$ be a set of formulas:
\begin{definition}\label{def:revision:representation}
  Let $B=(K,U,V)$ be a belief change scenario in ${\cal L}_{{\cal P}}$
  and let $(EQ_i)_{i\in I}$ be the family of all sets of equivalences,
  as defined in Definition~\ref{def:extension}.
  
  Define
  \(
  \lceil B\rceil
  \)
  as
  \(
  \mbox{$\bigvee_{i\in I}$}\mbox{$\bigwedge_{(s\in K)}$} \lceil s \rceil_i
  \)
  and
  \(
  \lceil B\rceil^c
  \)
  as
  \(
  \mbox{$\bigwedge_{(s\in K)}$} \lceil s \rceil_k
  \)
  for selection function $c$ corresponding to $EQ_k$.
%
%
%
\end{definition}
For revision, we define 
\(
\lceil{(K,\{\alpha\},\emptyset)}\rceil\wedge\alpha
\)
as the finite representation of $K\revise\alpha$,
and analogously
\(
\lceil{(K,\{\alpha\},\emptyset)}\rceil^c\wedge\alpha
\)
as the finite representation of $K\reviseC\alpha$.

\begin{theorem}\label{thm:rev.rep}
  Let $K$ and $\alpha$ be formulas.
  Then, we have
  \begin{eqnarray*}
  \theory{K}\revise\alpha
  & = &
  \theory{\;\lceil{(\theory{K},\{\alpha\},\emptyset)}\rceil\;\wedge \alpha}
  \\
  & \equiv &
  \lceil{(K,\{\alpha\},\emptyset)}\rceil \wedge \alpha.
  \end{eqnarray*}
\end{theorem}
Consider example (\ref{eq:ex:rev:tri}):
\(
\{p \wedge q\}\revise(\neg p \vee\neg q).
\)
So
\(
B={(\{p \wedge q\},\{(\neg p \vee\neg q)\},\emptyset)}
\)
is the belief change scenario.
We obtain:
\[
\!\!\!\!\!\!\!\!
\lceil B \rceil
\wedge(\neg p \vee\neg q)
=
[
(p \wedge\neg q)
\vee
(\neg p \wedge q)
]
\wedge
(\neg p \vee\neg q),
\]
which is equivalent to $(p \equiv\neg q)$.
For the other examples in Table~\ref{fig:revision.egs}, if $K$ is the
formula corresponding to $K'$ in the first column, then revising by the
given $\alpha$ via Theorem~\ref{thm:rev.rep}
is the formula given in the last line (up to permutation of symbols and
elimination of definitional equivalents).

Contraction is handled somewhat differently.
This is not altogether surprising, given that revision and contraction are
not fully interdefinable (Theorem~\ref{thm:Harper}).
Whereas for revision we replaced each atomic proposition in
$\overline{EQ_i}$ by its negation in $K$, for contraction replacements in
$K$ are done over all truth values of atomic propositions in
$\overline{EQ_i}$.
Formally, given a belief change scenario $B$, a corresponding set of
equivalences $EQ_i$ (according to Definition~\ref{def:extension})
along with its induced partition
\(
\langle{\cal P}_{EQ_i},{\cal P}_{\overline{EQ_i}}\rangle
\)
of ${\cal P}$, and a function
\(
\pi_{k_i}:{\cal P}_{\overline{EQ_i}}\rightarrow\{\top,\bot\}
\),
we define for
\(
\alpha
\in
{\cal L}_{{\cal P}}
\),
\(
\lfloor\alpha\rfloor^{k_i}
\)
as the result of replacing in $\alpha$ each proposition
\(
p\in{\cal P}_{\overline{EQ_i}}
\)
by $\pi_{k_i}(p)$.
Note that each set of equivalences induces a whole set $\Pi_i$ of such
mappings $\pi_{k_i}$,
viz.\
\(
\Pi_i
=
\{\pi_{k_i}\mid\pi_{k_i}:{\cal P}_{\overline{EQ_i}}\rightarrow\{\top,\bot\}\}
\),
amounting to all possible truth assignments to ${\cal P}_{\overline{EQ_i}}$.

\begin{definition}
  Let $B$ and $(EQ_i)_{i\in I}$ be defined as in
  Definition~\ref{def:revision:representation}.
  
  Define
  \(
  \lfloor B\rfloor
  \)
  as
  \(
  \mbox{$\bigvee_{i\in I,\pi_j\in\Pi_i}$}\mbox{$\bigwedge_{(s\in K)}$} 
  \lfloor s \rfloor^j
  \)
  and
  \(
  \lfloor B\rfloor^c
  \)
  as
  \(
  \mbox{$\bigvee_{\pi_j\in\Pi_k}$}\mbox{$\bigwedge_{(s\in K)}$}
  \lfloor s \rfloor^j
  \)
  for some selection function $c$ with $c(I)=k$.
%
%
\end{definition}
We define 
\(
\lfloor {(K,\emptyset,\{\neg\alpha\})}\rfloor
\)
as the finite representation of $K\contract\alpha$,
and analogously
\(
\lfloor {(K,\emptyset,\{\neg\alpha\})}\rfloor^c
\)
as the finite representation of $K\contractC\alpha$.
\begin{theorem}
\label{thm:con.rep}
  Let $K$ and $\alpha$ be formulas.
  Then, we have
  \begin{eqnarray*}
  \theory{K}\contract\alpha
  & = &
  \theory{\;\lfloor {(\theory{K},\emptyset,\{\neg\alpha\})}\rfloor\;}
  \\
  & \equiv &
  \lfloor {(K,\emptyset,\{\neg\alpha\})}\rfloor \ .
  \end{eqnarray*}
\end{theorem}
Consider example (\ref{eq:ex:con:one}):
\(
\{p \wedge\neg q\}\contract{(\neg q)}.
\)
We obtain
\[
\lfloor \;{(\{p \wedge\neg q\},\emptyset,\{q\})}\;\rfloor
=
(p \wedge\bot)\vee(p \wedge\top)
\equiv
p
\ .
\]
For the examples in Table~\ref{fig:contraction.egs}, if $K$ is the
formula corresponding to $K'$, then in contracting by the given $\alpha$,
the result of the contraction via Theorem~\ref{thm:con.rep}
is the formula given in the last line (up to permutation of symbols and
elimination of definitional equivalents).

Theorems~\ref{thm:rev.rep}~and~\ref{thm:con.rep} are interesting in that they
show that revision and contraction can be defined with respect to syntactic
objects (viz.\ sentences representing the knowledge base) yet are essentially
independent of syntactic form.
Hence in a certain sense the approach combines the advantages of base
revision \cite{Nebel92} and syntax-independent approaches.

\paragraph{Limiting the range of $EQ$:}

Intuitively, if an atomic sentence appears in a knowledge base $K$ but not in
a sentence for revision $\alpha$, or vice versa, then that atomic sentence
plays no part in the revision.
This is indeed the case here, as the next result demonstrates.
Let $\mathit{Vocab}(\delta)$ be the atomic sentences in $\delta$.
We obtain:
\begin{theorem}\label{thm:restrEQ}
Let $K$ be a set of formulas and $\alpha$ a formula.
Let
\(
E
=
\theory{K'\cup EQ\cup \alpha}\cap{\cal L}_{{\cal P}}
\)
be a consistent definitional extension of
belief change scenario $B=(\theory{K},\{\alpha\},\emptyset)$.

Then
\(
\{
p \equiv p' \mid
p \in
  (\mathit{Vocab}(K) \setminus \mathit{Vocab}(\alpha))
  \cup
  (\mathit{Vocab}(\alpha) \setminus \mathit{Vocab}(K))
\}
\subseteq
EQ.
\)
\end{theorem}
So for belief change, we need consider just the atomic sentences common to
$K$ and $\alpha$, and can ignore (with regards $EQ$) other atomic
sentences.
As detailed in the full paper, this result allows one to limit the
primed atomic propositions in $K'$ to those occurring in $\alpha$.

%
%
%
%
%
%
%
%
%
%

\section{The General Approach}
\label{sec:general}

Definition~\ref{def:extension} is quite general;
in Definitions~\ref{def:revision}~and~\ref{def:contraction} we narrow the
scope to specific approaches to belief change.
We note however, briefly, that other approaches are expressible in this
framework.
Belief {\em update} is a distinct form of belief change, suited to a
changing world.
Update and its dual operator {\em erasure} are studied in
\cite{KatsunoMendelzon92} where sets of postulates characterising the
operators are given.
\begin{definition}[Prime Implicate]
\label{def:prime.implicate}
A consistent set of literals $l$ is a {\em prime implicate}%
\footnote{Note that this is the dual of {\em prime implicant}.}
of $K$ iff:
$l \vdash K$ and for $l' \subset l$ we have $l' \not\vdash K$.
\end{definition}

\begin{definition}[Update]
\label{def:update}
Let $K$ be a knowledge base and $\alpha$ a formula and let
$PI(K)$ be the set of prime implicates of $K$.
For each $K_j \in PI(K)$, $1 \leq j \leq m$, let
\(
E^j_1, \dots, E^j_{n_j}
\)
be the consistent definitional extensions of
$(K_j,\{\alpha\}, \emptyset)$.
Then
\(
K\update\alpha
= \bigcup_{j=1}^{m} \bigcap_{i=1}^{n_j} E^j_i
\)
\quad
is the {\em update} of $K$ by $\alpha$.
\end{definition}
We do not define {\em choice update} here, given space limitations.
%
\begin{theorem}
$K\update\alpha$ satisfies the update postulates of
\cite{KatsunoMendelzon92}.
\end{theorem}
We show in the full paper that the operator $\update$ provides a
syntactic counterpart for Winslett's update operator \cite{Winslett88}.
We can also take a different notion of {\em maximal} in
Definition~\ref{def:extension}, and base the definition on set cardinality,
rather than set containment.
We show that based on this measure we can capture the revision approaches of
\cite{Dalal88} and \cite{Forbus89}.
Lastly a minor modification to Definition~\ref{def:extension} allows one to
use the framework to capture the {\em merging} of knowledge bases.

\section{Conclusion}

We have presented a general consistency-based framework for belief change,
having the same flavour as the consistency-based paradigms for diagnosis
or default reasoning.
We focus on a specific approach, in which a knowledge base $K$ and
sentence $\alpha$ are expressed, via renaming propositions in $K$,
in separate alphabets.
Given this, we assume that as many corresponding atoms in each language are
equivalent insofar as is consistently possible.
Lastly, we express the resultant knowledge base in a single language.
For the revision of $K$ by $\alpha$, for example, we begin with 
$\alpha$ and incorporate as much of $K$ as consistently possible. 
This gives rise to two notions of revision: 
a choice notion, in which one such ``extension'' is used for the revised 
state, and the intersection of all such extensions. 

The approach is well-suited for implementation:
The notion of a choice extension gives better complexity results than
general revision;
also, belief revision is an area in which choice reasoning may be useful.
Second, we show how the approach can be expressed in terms of a finite
knowledge base, and that the scope of a revision can be restricted
to those propositions common to the knowledge base and sentence for revision.

The approach allows for a uniform treatment of integrity constraints,
in that belief change may take into account both consistency-based and
entail\-ment-based static constraints, as well as dynamic constraints.
As well, it supports iterated revision.
Finally, the framework is applicable to other approaches to belief change.


\bibliographystyle{aaai}
\bibliography{ai}

\end{document}